\documentclass[journal]{IEEEtran}
\ifCLASSINFOpdf
\else
\fi
\usepackage{amsthm}
\usepackage{graphicx}
\usepackage{amsfonts}
\begin{document}
\copyright 2011 IEEE.  Personal use of this material is permitted.  Permission from IEEE must be obtained for all other uses, in any current or future media, including reprinting/republishing this material for advertising or promotional purposes, creating new collective works, for resale or redistribution to servers or lists, or reuse of any copyrighted component of this work in other works.”
\clearpage
\setcounter{page}{1}
%
\title{Towards Semantic Interoperability of Electronic Health Records}
%
%
%

\author{Idoia~Berges,
        Jes\'us~Berm\'udez
        and~Arantza~Illarramendi
\thanks{All authors are with the Department
of Languages and Information Systems, University of the Basque Country, Donostia-San Sebasti\'an, 20018 Spain. e-mail: \{idoia.berges, jesus.bermudez, a.illarramendi\}@ehu.es.}
\thanks{Manuscript received June 6, 2011.}}

\maketitle

\begin{abstract}
Although the goal of achieving semantic interoperability of Electronic Health Records (EHRs) is pursued by many researchers, it has not been accomplished yet. In this paper we present a proposal that smoothes out the way towards the achievement of that goal. In particular our work focuses on medical diagnoses statements. In summary the main contributions of our ontology-based proposal are the following: First, it includes a canonical ontology whose EHR-related terms focus on semantic aspects. As a result, their descriptions are independent of languages and technology aspects used in different organizations to represent EHRs. Moreover, those terms are related to their corresponding codes in well-known medical terminologies. Second, it deals with modules that allow obtaining rich ontological representations of EHR information managed by proprietary models of health information systems. The features of one specific module are shown as reference. Third, it considers the necessary mapping axioms between ontological terms enhanced with so-called path mappings. That feature smoothes out structural differences between heterogeneous EHR representations, allowing proper alignment of information. 
\end{abstract}


%
\IEEEpeerreviewmaketitle

\section{Introduction}

\IEEEPARstart{I}{n} 2009 the European Community presented a long-term research and
deployment roadmap that provides the key steps for achieving semantic interoperability in the area of healthcare\cite{Kalra09}. 
The incorporation some years ago of Electronic Health Records to the healthcare institutions may be seen as the first step towards the achievement of the goal, since, apart from local advantages over manual records such as avoiding legibility problems, they favour a fast exchange of clinical data between different organizations. However, the fact that most healthcare institutions have developed their health information systems in an autonomous way has resulted in a proliferation of heterogeneous health information systems, each one with its own proprietary model for representing and storing EHR information, which hinders the task of interoperating with each other. 

In many areas, the adoption of knowledge representation standards stands out as the most usual approach to solve interoperability problems. This happens also in the healthcare area, where some standards such as openEHR\cite{openEHRsoloURL}, ISO 13606\cite{ISO13606} and HL7-CDA\cite{HL7-CDAsoloURL} are under development for this purpose. 
All three follow a dual model-based methodology for representing EHR information: the Reference Model defines basic structures such as List, Table, etc., while the Archetype Model defines knowledge elements (such as Respiration Rate) by using and constraining the elements of the Reference Model. 
Although the idea of using a standard may seem suitable for the considered goal, we think that interoperability does not mean to have a unique representation but a semantically acknowledgeable equivalent one. This would relieve healthcare institutions from being forced to use one standard in the representation of their knowledge and moreover, since several standards are being developed for the same purpose, the interoperability problem will remain unsolved unless these standards merge into a single one. Currently, some research is being done on the latter issue\cite{Schloeffel06}.

In this paper we present a proposal to move towards the notion of full semantic interoperability of heterogeneous EHRs, which states that when one particular system receives some EHR information from another healthcare institution, the received information can be seamlessly integrated into its underlying repository because the differences in the language, in the representation of the information and in the storing systems do not cause any misunderstanding\cite{Kalra09}.
Two general approaches for interoperability among systems are described in \cite{Kashyap96}: Using a canonical model to which the particular models are linked or aligning the particular models two by two. The proposal presented in this paper is sustained in the former approach. More precisely, it is an ontology-based approach where OWL2\cite{OWL2overview} ontologies are used as representation models. In general, ontologies have been considered relevant for several purposes such as: enabling reuse of domain knowledge, allowing the analysis of domain knowledge and sharing common understanding of the meaning of information\cite{Uschold96}. Our approach benefits from the latter advantage and additionally it provides the following ones:

\begin{itemize}
\item It favors the notion of semantic interoperability: The use of a formal ontology as canonical conceptual model allows to focus on  aspects that are independent of the languages or technologies used to describe EHRs.
\item It favors the notion of extensibility to different models: The framework comprises two kinds of ontologies which represent the definitions of clinical terms that appear in EHRs at different levels of abstraction. The canonical contains ontological definitions of EHR statements and the application ontologies contain specializations of the definitions of the canonical ontology according to the standards mentioned previously or according to proprietary models of healthcare institutions.
\item It decreases the need of human intervention: The framework relies on a reasoning mechanism that, using axioms stated in the ontology, infers knowledge that allows the discovery of more relationships among the heterogeneous models used by the different health information systems.
\end{itemize}

Dealing with ontologies, one relevant aspect is the features of the terms that are part of them. In our scenario those terms are related to EHRs.
Different kinds of information can be found in an EHR. OpenEHR divides this information into 5 subtypes\cite{Beale07} and we also have adopted that division in the definition of our canonical ontology: \textit{Observations} comprise the data that can be measured in an objective way, such as the age of a patient, his respiration rate, etc. \textit{Evaluations} represent the evidence obtained from observations, for example the diagnosis of an illness. \textit{Instructions} represent actions to be performed in the future such as the prescription of a medicine or the request of a laboratory test. \textit{Actions} are used to model the information recorded due to the execution of an instruction and finally there is one last type to record \textit{administrative} events such as admission or discharge information. In this paper we just focus on one type of evaluations, namely the diagnoses, but similar ideas to those that will be explained here for diagnoses could be also applied to the other types of information. Moreover, the terms of the application ontologies are obtained from the particular health information systems and then linked to the terms in the canonical ontology by using ontology mappings.

A certain number of works related to ours can be found at present. With regard to the benefits of taking semantics into account, some works discuss the convenience of using semantic technologies in several heathcare related issues. In \cite{Wroe06}, the handicaps for widespread adoption of semantic technologies within a care records system are pinpointed. In \cite{WeberJahnke10} the challenges to be addressed in order
to be able to use the so-called Smart Internet to enable reforms on healthcare information systems are discussed. Lastly, in \cite{Krummenacher09} the triplespace paradigm is suggested as semantic middleware to support pervasive access to
electronic patient summaries.
The works mentioned next also rely on semantic technologies for interchanging data, as opposed to other formats such
as XML, which are structure-based. More specifically, related to the topic of facilitating semantic interoperability between heterogeneous health information systems, the following works deal only with the interoperability between standard-based health information systems: \cite{Kilic09}
provides a solution to achieve semantic interoperability between systems that have been developed under the
HL7 reference model and which requires that the source system has some prior knowledge about the target
system. In \cite{Bicer05} ontology mappings are proposed between pairs of archetype-based
models. 
In \cite{MartinezCosta10} a model-driven engineering approach that transforms archetypes of the ISO 13606 standard into OWL models is presented. Finally, authors in \cite{Lezcano11} describe an approach to translate from the Archetype Definition Language (ADL\cite{ADL}) to OWL, they also present some techniques to map archetypes to formal ontologies and show the convenience of using semantic rules on the resulting representation in order to guide the execution of primary care guidelines. In this paper we present a wider approach since apart from the interoperability of standard-based systems we deal also with interoperability considering proprietary models. Some other works that tackle the problem of semantic interoperability of EHRs from a different perspective are the following: In \cite{PradosSuarez11} a semantic conceptualization model for an EHR system is presented. This still early work is more oriented towards the accessibility, use and management of the EHR at a local level, but it also aims at providing a base in order to solve the interoperability problem from a semantic point of view.
In \cite{Hedayat10} the hypothesis that semantic technologies are potential bridging technologies between the EHRs and medical terminologies --as well as a possible representation of the combined semantics of systems to be integrated-- is raised and some experimental study is made on this issue. We also promote the connection between the semantic representation of EHR statements and their codes in well-known terminologies. Finally, in \cite{Gonzalez10} authors discuss how advanced middleware, such as Enterprise Middleware Bus, and semantic web services can assist in solving interoperability issues between eHealth systems.

The rest of the paper is divided as it follows: In Section \ref{sec:framework} the global architecture of the framework is presented, and extensive details about the canonical ontology and the auxiliary modules DB2OntoModule and MappingModule are given. The feasibility of the solution is shown in Section \ref{sec:scenario}, and finally, conclusions are discussed in Section \ref{sec:conclusions}.

\section{Global architecture}\label{sec:framework}
\begin{figure}
\begin{center}	
	\includegraphics[width=3.5in]{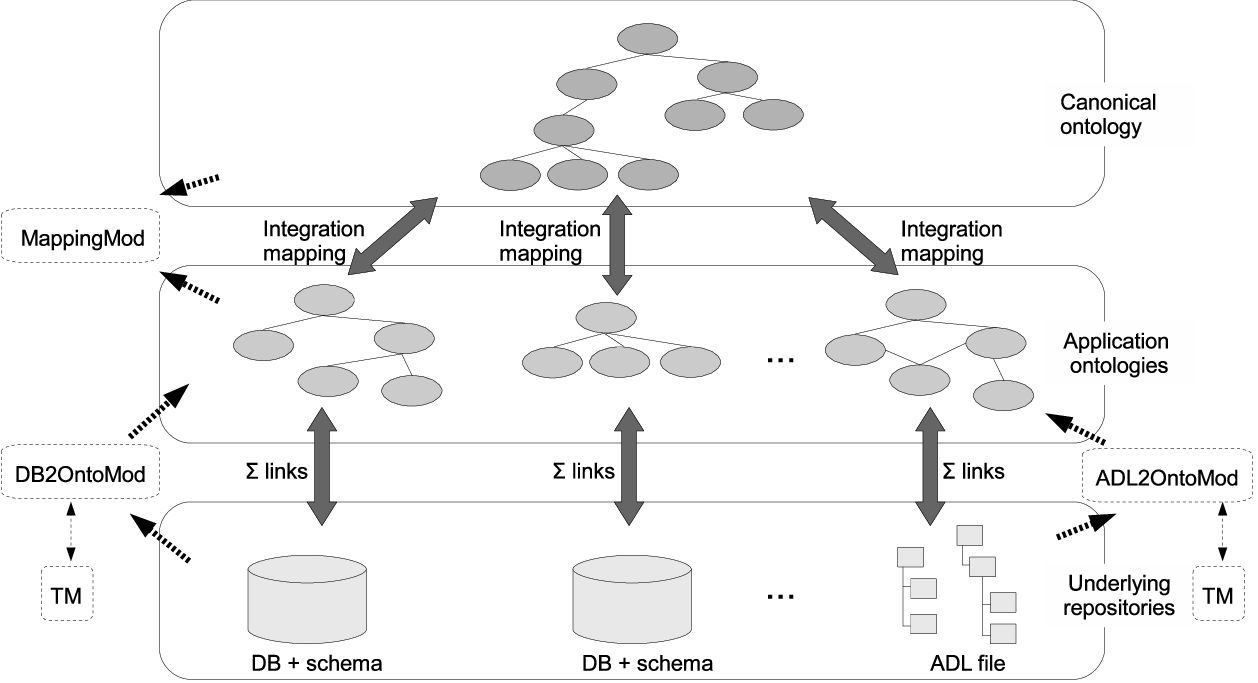}
	\caption{\label{architecture} Global architecture of the solution}
\end{center}
\end{figure}

In Fig. \ref{architecture} the three-layered architecture of the solution can be found. The \textit{lower layer}, contains the particular underlying repository of each healthcare institution, where the information of the EHRs is stored. Associated to each kind of underlying repository, there is some kind of file (e.g. database schema, set of ADL files) where information about the structures in the repository can be found. Then, the \textit{middle layer} contains one application ontology for each information system, built on top of the underlying repository. These application ontologies are created semi-automatically from the underlying repositories by some auxiliary modules (e.g. DB2OntoModule and ADL2OntoModule), or imported from an ontology repository, and describe semantically each underlying repository. Moreover, they are linked to their corresponding repositories by some $\Sigma$ links that specify how to transfer information from each of the representations to the other. Finally the \textit{upper layer} contains one canonical ontology. This ontology will contain the necessary classes and properties to represent the different types of information that can be found in an EHR and is linked to each of the application ontologies by some integration mappings defined by a MappingModule. Each particular healthcare institution will have only a partial view of the global framework, since with our proposal there is no need for that institution to know anything but its underlying repository, its application ontology, and the canonical ontology. 

The proposed framework allows one healthcare institution to interpret on the fly clinical statements sent by another one  --even when they use proprietary formats. We support our claim on the following techniques:

\begin{itemize}
\item {\it Logic-based descriptions:} Representations of diagnoses considered by particular health information systems, described using standards as well as proprietary models, are expressed in our approach by using OWL2 ontology axioms. Moreover, terms in those axioms are related with canonical ontology terms that focus their  descriptions on 
language and technology independent aspects. 
This approach increases the opportunities of solving the interoperability issue since it relies mainly on semantic aspects.
\item {\it Automated reasoning:} All ontology descriptions, as well as the mappings among elements of 
the ontologies, are expressed in the same formalism OWL2. This uniform representation allows the 
use of well-known reasoners in order to derive new statements from the existing ones. Furthermore, 
the mismatch problem is avoided and automatic integration is facilitated.
\item {\it Transfer mechanism:} A process, guided by the previous two items, is implemented to 
transform a particular clinical statement from a healthcare institution into a corresponding 
clinical statement for another healthcare institution. So-called path mappings play a crucial role during the transfer process, smoothing out the structural differences between EHR representations.
\end{itemize}

Finally, we are aware that the messiness of real world EHRs may sometimes hinder the task of fitting them into the presented proposal, but in our opinion this does not invalidate the advantages it can provide in many situations. 

In the following subsections, the canonical ontology, the \textit{DB2OntoModule} and the \textit{MappingModule} are described thoroughly.

\subsection{Canonical ontology: Representing diagnoses in OWL}\label{sec:canonical}
\newtheorem{definition}{Definition}

The canonical ontology contains the necessary classes and properties to represent the different types of information that can be found in an EHR. Following openEHRs classification of EHR entries, we have defined in the ontology five classes to represent the general categories: \footnotesize \texttt{Observation}, \texttt{Evaluation}, \texttt{Instruction}, \texttt{Action} \normalsize and \footnotesize \texttt{Admin}. \normalsize Moreover, these five classes have been specialized to represent more specific types of entries. As we pointed out in the introduction, in this paper we deal with diagnoses, which are a special case of evaluations.


A \textit{diagnosis} is defined as the act of identifying a disease from its signs and symptoms, as well as the decision reached by that act\footnote{http://www.merriam-webster.com/dictionary/diagnosis}. For this reason, in addition to representing a diagnosis as a subclass of \footnotesize \texttt{Evaluation}, \normalsize its definition is enhanced with two properties: \footnotesize \texttt{hasFinding}, \normalsize to indicate the conclusion reached by the physician about what is happening to the patient, and \footnotesize \texttt{hasObs}, \normalsize to indicate the information about the observation(s) which lead to that assessment\footnote{For the presentation we prefer a logic notation instead of the more verbose \textsc{rdf/\textsc{xml}} syntax.}.

\scriptsize
\begin{eqnarray*}
\texttt{Diagnosis}&\equiv&\texttt{Evaluation}\sqcap \texttt{=1 hasFinding.Finding}\sqcap\\
&&\exists\texttt{hasObs.Observation}\\
\end{eqnarray*}
\normalsize
Specific diagnoses are defined as subclasses of the class \footnotesize \texttt{Diagnosis}. \normalsize For example, the evidence obtained as a result of an ECG can be described by specializing the range restrictions of the properties \footnotesize \texttt{hasFinding} \normalsize and \footnotesize \texttt{hasObs}. \normalsize For instance, the observation that leads to an ECG diagnosis is an ECG Recording, which is made up of several components\footnote{For the sake of brevity, in this example only some components of an ECG are taken into account. Please refer to \cite{openEHRsoloURL} for the whole set of components.}: some of the components refer to information about the heart's electrical axis (i.e. the general direction of the heart's depolarization wavefront), while the others refer to information about the entire ECG.

\scriptsize
\begin{eqnarray*}
\texttt{ECGDiagnosis}&\equiv&\texttt{Diagnosis}\sqcap \texttt{=1 hasFinding.ECGFinding}\sqcap\\
&&\exists\texttt{hasObs.ECGRecording}\\
\texttt{ECGRecording}&\equiv&\texttt{Observation}\sqcap\exists\texttt{comp.P-Axis}\sqcap\\
&&\exists\texttt{comp.QRS-Axis}\sqcap\exists\texttt{comp.T-Axis}\sqcap\\
&&\exists\texttt{comp.PR-Interval}\sqcap\exists\texttt{comp.QT-Interval}\sqcap\\
&&\exists\texttt{comp.QTc-Interval}\sqcap\exists\texttt{comp.QRS-Duration}\sqcap\\
&&\exists\texttt{comp.Heart-Rate}
\end{eqnarray*}
\normalsize

One advantage of working in the medical area is the existence of medical terminologies, such as SNOMED\cite{SNOMEDsoloURL} and LOINC\cite{LOINCsoloURL}. These terminologies cover most areas of clinical information and provide a consistent way to identify medical terms univocally, which can be very helpful at the time of gathering and exchanging clinical results. Our system takes advantage of these terminologies to enhance the definition of the classes in the canonical ontology. Thus, whenever is possible, each term in the ontology is related to its corresponding code in those terminologies: 

\scriptsize
\begin{eqnarray*}
\texttt{ECGDiagnosis}&\equiv&\exists\texttt{loinc.\{`8601-7'\}}\\
\texttt{ECGRecording}&\equiv&\exists\texttt{loinc.\{`34534-8'\}}\\
\texttt{P-Axis}&\equiv&\exists\texttt{loinc.\{`8626-4'\}}
\end{eqnarray*}
\normalsize

The use of terminological codes into the definitions of the classes in the ontology increases the chances of achieving a successful communication. 

Finally, since building a canonical ontology is not an easy task, we think that efforts that are being done to define archetypes in openEHR could be reused to achieve that task.

\subsection{DB2OntoModule}\label{sec:db2onto}
Taking into account the widespread use of relational databases to store EHR records, we show in this subsection the main features of the module DB2OntoModule\footnote{Other modules, such as the ADL2Onto module, would be used to perform the translations between other sources and the ontology}. This module takes as input a database schema and after applying a set of rules based on schema features, it obtains the ontological representations of those relational databases (i.e. the application ontology of that system). 
In the specialized literature many approaches for translating relational structures into more expressive formalisms can be found: object models, description logics and Semantic Web technologies. Some of them follow the so-called declarative approach, which first convert the relational structure into a declarative language and then the result is modified by the user to declare additional knowledge about the database (e.g \cite{Champin07}). Our proposal also uses the declarative approach but its novel contribution relies in the large number of schema properties that it considers, allowing to make explicit more knowledge, and in the fact that it associates to the obtained classes their corresponding codes that appear in well-known medical terminologies. 

In order for the DB2OntoModule to accomplish the last feature, it deals with an element called ``Terminology Manager'' or, in short, ``TM'', which has an associated function of the form \textit{getX(conceptName)}, where \textit{X} is the name of a terminology (LOINC, SNOMED, or any other) and \textit{conceptName} is the name of the relation or attribute whose terminological code is to be retrieved. For example, in the case of a relation \textit{BloodPressure(id, systolic, diastolic)} the TM would contain:
\vspace{0.2cm}

\small
\begin{center}
\begin{tabular*}{2.65in}{|l|l|l|}
\hline
\textbf{Identifying path}&\textbf{LOINC}&\textbf{SNOMED}\\
\hline
\textit{BloodPressure}&18684-1&75367002\\
\hline
\textit{BloodPressure.systolic}&8480-6&72313002\\
\hline
\textit{BloodPressure.diastolic}&8462-4&271650006\\
\hline
\end{tabular*}
\end{center}
\normalsize

\vspace{0.2cm}
Concerning schema features, the DB2OntoModule works as follows:

\noindent\textbf{Relations:} Relations of the relational schema are translated into OWL2 classes. Moreover, if for a given relation $R$, TM.\textit{getLOINC($R$)}=$LC$ (being $LC$ a particular LOINC code), a new axiom \texttt{R}$\equiv\exists$\texttt{loinc.\{'$LC$'\}} is added to the ontology (analogously for other terminological codes).

\noindent\textbf{Attributes:} 
Two options arise:
(1) If for a given attribute $a$ in $R$ ($R.a$) TM.\textit{getLOINC($R.a$)} returns some code $LC$, then a new class \texttt{A} is created (if there is no other class which already has that code). Moreover, the axioms \texttt{A}$\equiv\exists$\texttt{loinc.\{'$LC$'\}} and  \texttt{A}$\sqsubseteq\exists$\texttt{value.}\textit{getType($a$)} are added. Finally, if attribute $a$ is compulsory in $R$, the axiom  \texttt{R}$\sqsubseteq\exists$\texttt{hasA.A} is added.
(2) If there is no code for $R.a$ in TM, a property \texttt{a} is created in the ontology, where \textit{Domain(\texttt{a})}=\texttt{R} and \textit{Range(\texttt{a})}=\textit{getType($a$)}. Moreover, if attribute $a$ is compulsory in $R$, the axiom \texttt{R}$\sqsubseteq\exists$\texttt{a} is added.

\noindent\textbf{Integrity constraints:} An integrity constraint such as \textit{R.a}$>$30 adds a new axiom of type \texttt{R}$\sqsubseteq\exists$
\texttt{hasA.(A and $\exists$value[>30])} if 
$R.a$ is in TM, and a new axiom of type \texttt{R}$\sqsubseteq\exists$ \texttt{a[>30]} otherwise.

Once the previous steps are accomplished, the next one involves enriching the obtained descriptions by using several types of information, such as inclusion, exclusion and functional dependencies:

\noindent\textbf{Inclusion dependencies:} Three different situations are considered (see a previous work \cite{Blanco94} from our group for more details): 
(1) Dependencies between key (\textit{R.K}) and non-key (\textit{S.x}) attributes, which indicate the existence of a foreign keys of type \textit{S.x}$\subseteq$\textit{R.K}. These dependencies are reflected by defining an association between the ontology classes obtained from those relations (\texttt{S}$\sqsubseteq\exists$\texttt{x.R}); (2) Dependencies between the keys of two relations (\textit{R.K}$\subseteq$\textit{R'.K'}); and (3) Dependencies between a subset of a key and a key (\textit{R.subK}$\subseteq$\textit{R'.K'}), which also have the corresponding reflection.

\noindent\textbf{Exclusion dependencies:} An exclusion dependency between the keys of two relations (\textit{R.K}$\cap$\textit{R'.K'}=$\emptyset$) creates a new axiom of the form \texttt{R}$\sqsubseteq\neg$\texttt{R'} in the ontology. In addition, if there is no class in the ontology that subsumes both \texttt{R} and \texttt{R'}, such new class \texttt{S} is created and the axioms \texttt{R}$\sqsubseteq$\texttt{S} and \texttt{R'}$\sqsubseteq$\texttt{S} are added.

\noindent\textbf{Functional dependencies:} 
If a functional dependency of the form \textit{R.X}$\rightarrow$ \textit{R.y} is detected, with \textit{X} and \textit{y} being a non-key attribute set and a non-key attribute respectively, a new class \texttt{X} is created. Moreover, two  new properties \texttt{hasX} and \texttt{hasY} are defined and the axioms \texttt{R}$\sqsubseteq\exists$\texttt{hasX.X} and \texttt{X}$\sqsubseteq\exists$\texttt{hasY}.\textit{getType(y)} are added to the ontology. 

Furthermore, the ontology can be enriched by using domain information for attribute values, for example, in the case of properties expressed by enumerating attribute values. For an attribute $R.a$ whose possible values are either $A1$ or $A2$, if both have a corresponding code in the TM,  classes \texttt{A1} and \texttt{A2} are created in the ontology. Moreover, one general class to group those two classes is created (e.g. \texttt{A0}) and axioms \texttt{A1}$\sqsubseteq$\texttt{A0}, \texttt{A2}$\sqsubseteq$\texttt{A0}, \texttt{A1}$\sqsubseteq$$\neg$\texttt{A2} and \texttt{R}$\sqsubseteq\forall$\texttt{a.A0} are added. However, in the case where $A1$ and $A2$ have no terminological code in the TM, class \texttt{A0} is created as an enumeration of two individuals $a1$ and $a2$, and axiom \texttt{R}$\sqsubseteq\forall$\texttt{a.A0} is added too.

All the previous types of considerations are applied in the following sequence: first inclusion dependencies; then when the input relational schema is not in second or third normal form, functional dependencies are used to create new classes; next exclusion dependencies are exploited and last integrity constraints and domain information for attribute values are considered.
Finally, once the DB2OntoModule has performed the steps above, a candidate ontology has been created. However, we feel that it is advisable to allow the health system administrator to modify the ontology in a flexible way. For example, some common changes could be substituting $\sqsubseteq$ relationships with $\equiv$ relationships, modifying the names of the terms that have been created, or adding some missing terminological code. These changes can be done manually using any well-known ontology editor.

\subsubsection*{The DB2OntoModule at work}
For example, a particular registration for an ECG diagnosis may consist of four relational tables according to the following schema (all attributes are considered compulsory)

\scriptsize
\begin{eqnarray*}
& & \texttt{ECGDiagnosis(\underline{code}, finding, recording)}\\
& & \texttt{ECGObservation(\underline{code}, axis, global)}\\
& & \texttt{ECGAxis(\underline{code}, P-Axis, QRS-Axis, T-Axis)}\\
& & \texttt{ECGGlobal(\underline{code}, PR-Interval, QT-Interval,}\\& &\ \ \ \ \ \texttt{QTC-Interval, QRS-Duration, Heart-Rate)}\\
\end{eqnarray*}
\normalsize 
and the following inclusion dependencies between non-key and key attributes:

\scriptsize
\begin{eqnarray*}
& & \texttt{ECGDiagnosis.recording} \subseteq \texttt{ECGObservation.\underline{code}}\\
& & \texttt{ECGObservation.axis} \subseteq \texttt{ECGAxis.\underline{code}}\\
& & \texttt{ECGObservation.global} \subseteq \texttt{ECGGlobal.\underline{code}}\\
\end{eqnarray*}
\normalsize

Moreover, let us consider the bogus case where the attribute \footnotesize \texttt{finding} \normalsize of the ECGDiagnosis table must be either ``Normal ECG'' or ``Abnormal ECG''. As a result of applying the initial steps for transforming the schema to ontology elements four new classes are created in the ontology: \footnotesize \texttt{EGCDiagnosis}, \texttt{ECGObservation}, \texttt{ECGAxis} \normalsize and \footnotesize \texttt{ECGGlobal}, \normalsize each with its respective LOINC code. 
Moreover, since the compulsory attribute \footnotesize \texttt{P-Axis}\normalsize, whose type is ``int'', has also a LOINC code at the TM, axioms \footnotesize \texttt{P-Axis}$\equiv\exists$\texttt{loinc.\{`8626-4'\}}, \texttt{ECGAxis}$\sqsubseteq\exists$\texttt{hasP-Axis.P-Axis} \normalsize and \footnotesize \texttt{P-Axis}$\sqsubseteq\exists$\texttt{value.xsd:int} \normalsize are created (same process for the other attributes). Then, the rules for inclusion dependencies are applied, and, for example, from the inclusion dependency \footnotesize \texttt{ECGObservation.axis} $\subseteq$ \texttt{ECGAxis.\underline{code}}, \normalsize axiom \footnotesize \texttt{ECGObservation}$\sqsubseteq\exists$\texttt{axis.ECGAxis} \normalsize is created. Moreover, information about the allowed values for the \footnotesize \texttt{finding} \normalsize attribute is considered and a new class \footnotesize \texttt{ECGFinding} \normalsize is created as superclass of two other classes \footnotesize \texttt{NormalECG} \normalsize and \footnotesize \texttt{AbnormalECG}.
\normalsize
Finally, manual changes are applied. For example, we have chosen to substitute some of the subclass relationships with equivalence relationships, so the created ontology has, among others, the following axioms\footnote{Throughout the paper, namespaces \texttt{a:} and \texttt{b:} will be used to refer to terms in the application ontologies of two particular systems $A$ and $B$. Moreover, namespace \texttt{c:} or no namespace are used to indicate the terms in the canonical ontology.}:

\scriptsize
\begin{eqnarray*}
\texttt{a:ECGDiagnosis}&\equiv&\exists \texttt{a:finding.a:ECGFinding}\sqcap\\
&&\exists\texttt{a:recording.a:ECGObservation}\\
\texttt{a:ECGDiagnosis}&\equiv&\exists\texttt{a:loinc.\{8601-7\}}\\
\texttt{a:ECGObservation}&\equiv&\exists\texttt{a:hasAxis.a:ECGAxis}\sqcap\\
&&\exists\texttt{a:hasGlobal.a:ECGGlobal}\\
\texttt{a:ECGObservation}&\equiv&\texttt{$\exists$loinc.\{34534-8\}}\\
\texttt{a:ECGAxis}&\equiv&\exists\texttt{a:hasP-Axis.a:P-Axis}\sqcap\\
&&\exists\texttt{a:hasQRS-Axis.a:QRS-Axis}\sqcap\\
&&\exists\texttt{a:hasT-Axis.a:T-Axis}\\
\texttt{a:NormalECG}&\equiv&\texttt{a:ECGFinding}\sqcap\texttt{a:value.\{"Normal ECG"\}}\\
\texttt{a:NormalECG}&\equiv&\texttt{a:snomed.\{102593009\}}
\end{eqnarray*}
\normalsize

The second task of the DB2OntoModule is to create the $\Sigma$ links that indicate how to transfer the information from the database to the ontology that has been created from it (and vice versa). This task was previously tackled by our research group, so we refer to the reader to \cite{Blanco99} for further technical details. 

\subsection{MappingModule}\label{sec:mappingmodule}

Once an application ontology of one particular system has been generated by the corresponding translator module, it must be integrated with the canonical ontology, and the mappings between the terms of that application ontology and the canonical ontology must be created. A \textit{MappingModule} has been implemented for this purpose. Wide research has been done in the specialized literature about ontology mapping (e.g. \cite{Euzenat07}), so working in new techniques for that same issue is out of the scope of our work. So, our MappingModule takes a pragmatic approach and receives as input a set of basic mapping axioms specifically defined by the system administrator (for example, to state that the property \texttt{a:loinc} is equivalent to the property \texttt{c:loinc}). Then, it incorporates these basic mappings into the ontologies and, with the help of a reasoner, it creates an integration mapping that relates the terms of the application ontology with those of the canonical ontology. 

However, our module presents a distinguishing feature, since it considers mappings between ontology paths, which are rarely considered in other works. In order to be aware of the importance of discovering path mappings, let us compare the definitions of classes \footnotesize\texttt{c:ECGRecording} \normalsize and \footnotesize\texttt{a:ECGObservation} \normalsize in sections \ref{sec:canonical} and \ref{sec:db2onto} respectively. Both share the same LOINC code (34534-8), so their semantics are the same. 
Looking at the description of \footnotesize\texttt{c:ECGRecording} \normalsize, it can be seen that any individual belonging to that class will be directly related to an individual of the class \footnotesize\texttt{c:P-Axis} \normalsize via the property \footnotesize\texttt{c:comp} \normalsize(assume the same intuition for the other components). However, in the case of the descriptions in the application ontology of system A, it turns out that classes \footnotesize\texttt{a:ECGObservation} \normalsize and \footnotesize\texttt{a:P-Axis} \normalsize are not directly related, but indirectly: first \footnotesize\texttt{a:ECGObservation} \normalsize is related to the class \footnotesize\texttt{a:ECGAxis} \normalsize via the property \footnotesize\texttt{a:hasAxis} \normalsize and then the class \footnotesize\texttt{a:ECGAxis} \normalsize is related to the class \footnotesize\texttt{a:P-Axis} \normalsize via the property \footnotesize\texttt{a:hasP-Axis} \normalsize. Then it could be stated that there is a simple path between classes \footnotesize\texttt{c:ECGRecording} \normalsize and \footnotesize\texttt{c:P-Axis} \normalsize, while there is a composite path between classes \footnotesize\texttt{a:ECGObservation} \normalsize and \footnotesize\texttt{a:P-Axis} \normalsize.

\normalsize
Intuitively, those two paths could be regarded as equivalent, since their only difference is from the structural point of view caused by the heterogeneous origin of the ontologies, not from a semantic point of view. Let us show how our module deals with that aspect:

\begin{definition}
An \textit{ontology path} is a regular expression of the form \texttt{A.}(\texttt{p.[B]})$^{+}$ where \texttt{A,B} represent class names and \texttt{p} represents property names, all from the same ontology.
\end{definition}

Let us denote equivalences between paths with the symbol $\equiv_{p}$. For instance, the aforementioned example is represented as:

\scriptsize
\begin{eqnarray*}
&\hspace{-3mm}\texttt{a:ECGObservation.a:hasAxis[a:ECGAxis].a:hasP-Axis.[a:P-Axis]} & \\
&\equiv_{p}&\\
&\texttt{c:ECGRecording.c:comp[c:P-Axis]}& 
\end{eqnarray*}
\normalsize

Although in this example an equivalence path mapping has been presented, a corresponding idea is valid for subclass path mappings ($\sqsubseteq_{p}$) and superclass path mappings ($\sqsupseteq_{p}$). In order to determine path mappings, first path mapping candidates are searched:

\newcommand{\cp}{\stackrel{\sqsubset}{\sim}}

\begin{definition}
Let $Path_{C}=C_{0}\textbf{.}p_{1}[C_{1}]...\textbf{.}p_{n}[C_{n}]$ and $Path_{D}=D_{0}\textbf{.}q_{1}[D_{1}]...\textbf{.}q_{m}[D_{m}]$ be two ontology paths. A path mapping candidate exists between $Path_{C}$ and $Path_{D}$ if any of the following statements holds:
\begin{itemize}
\item $C_{0}\sqsubseteq D_{0}$ and $C_{n}\sqsubseteq D_{m}$(represented as $Path_{C}\stackrel{\sqsubseteq}{\sim} Path_{D}$)
\item $C_{0}\sqsupseteq D_{0}$ and $C_{n}\sqsupseteq D_{m}$(represented as $Path_{C}\stackrel{\sqsupseteq}{\sim} Path_{D}$)
\end{itemize}
Moreover, if $Path_{C}\stackrel{\sqsubseteq}{\sim} Path_{D}$ and $Path_{C}\stackrel{\sqsupseteq}{\sim} Path_{D}$ then $Path_{C}\stackrel{\equiv}{\sim} Path_{D}$
\end{definition}

A path mapping candidate becomes a proper path mapping when the semantics of both paths is found to be the same.
Path mappings are useful at the time of transforming individuals from one ontology so that they meet the requirements of the target ontology. The implementation of path mappings is done by using SWRL\cite{SWRLsoloURL} rules. 
SWRL increases the expressivity of OWL and thus allows to model more domain knowledge than the one achieved by using OWL in its own. Moreover, since SWRL can be tightly integrated with OWL, there is no impedance
mismatch between the modelling language and the rules language: SWRL rules can use directly the classes, properties and individuals defined in the OWL model.
For example, the path mappings shown before would be implemented using the following rules (one in each way):

\vspace{0.2cm}

\footnotesize
\begin{description}
\item[$R1$] \texttt{a:ECGObservation(?e)} $\wedge$ \texttt{a:hasAxis(?e,?x)} $\wedge$ 
\texttt{a:hasP-Axis(?x,?p)} $\rightarrow$ \texttt{c:comp(?e,?p)} 
\item[]
\item[$R2$]  \texttt{c:ECGRecording(?e)} $\wedge$ \texttt{c:comp(?e,?p)} $\wedge$ \texttt{c:P-Axis(?p)}  $\wedge$ \texttt{swrlx:createOWLThing(?e,?x)} $\rightarrow$ \texttt{a:hasAxis(?e,?x)} $\wedge$ \texttt{a:ECGAxis(?x)} $\wedge$ \texttt{a:hasP-Axis(?x,?p)} 
\end{description}
\normalsize
\vspace{0.2cm}

As looking for all the candidate path mappings between two large ontologies might be a hard task considering both time and resources, a threshold can be established to indicate the maximum length of the paths to be searched. Some other heuristics could be applied too to discover candidate path mappings efficiently.

So, to sum up, the \textit{integration mapping} that is generated between an application ontology and the canonical ontology can be defined as it follows:

\begin{definition}\label{def:integrationm}
An \textit{integration mapping} is a structure 
$\mathcal{I} = \langle O, G, \mathcal{M} \rangle$ where $O$ is a set of OWL2 axioms that 
comprises the application ontology corresponding to a healthcare institution, $G$ is the 
set of OWL2 axioms for the canonical ontology, and $\mathcal{M}$ is a set of 
\textit{mapping axioms} of any of the following forms:
\begin{itemize}
\item $C_{o} \sqsubseteq Exp_{g}$, $C_{o} \sqsupseteq Exp_{g}$ or 
$C_{o} \equiv Exp_{g}$, where  
$C_{o}$ is a class name from $O$, and $Exp_{g}$ is a OWL2 class expression that uses only terms from $G$.\\
\item $p_{o} \sqsubseteq p_{g}$ or $p_{o} \sqsupseteq p_{g}$ , where  
$p_{o}$ is a property name from $O$, and $p_{g}$ is property name from $G$.\\
\item $sameAs(i_{o}, i_{g})$ , where  
$i_{o}$ is the name of an individual from $O$, and $i_{g}$ is the name of an individual from $G$.\\
\item $Path_{o} \sqsubseteq_{p} Path_{g}$, $Path_{o} \sqsupseteq_{p} Path_{g}$ or 
$Path_{o} \equiv_{p} Path_{g}$, where $Path_{o}$ is an ontology path in $O$ and $Path_{g}$ is an ontology path in $G$.
\end{itemize}
\end{definition}

The result of the engineering process of producing the set $\mathcal{M}$ of mapping axioms is the key for the interoperability of two different health information systems.

\section{Framework at work}\label{sec:scenario}

The main contribution of our proposal is the capability of one system $B$ of interpreting information sent by another system $A$ on the fly, without prior peer-to-peer agreement on the semantics and syntax of the interchanged data. In this example, let us suppose that the database schema of system $A$ is the one presented in section \ref{sec:db2onto}. Moreover, in the case of system $B$, let us consider that it follows the HL7 standard and that different representations are used to represent ECG information depending on the result of the ECG (e.g.: \footnotesize \texttt{ECGNormalDiag} \normalsize for normal ECG results, \footnotesize  \texttt{ECGAbnormalDiag} \normalsize when abnormalities have been detected). The work of the ADL2OntoModule and MappingModule led to the following axioms, with respect to the application ontology of system $B$: 

\scriptsize
\begin{eqnarray*}
\texttt{b:ECGNormalDiag}&\equiv&\texttt{b:ECGDiagnosis}\sqcap\\
&& \texttt{=1 b:finding.b:ECGNormalFind}\sqcap
\\
&&\exists\texttt{b:component.b:P-Ax}\sqcap\\
&&\exists\texttt{b:component.b:QRS-Ax}\sqcap\texttt{...}\sqcap\\
&&\exists\texttt{b:component.b:Heart-R}\\
\texttt{b:ECGDiagnosis}&\equiv&\texttt{b:loinc.\{8601-7\}}\\
\texttt{b:ECGNormalFind}&\equiv&\texttt{b:snomed.\{102593009\}}\\
\texttt{b:loinc}&\equiv&\texttt{c:loinc}\\
\texttt{b:component}&\equiv&\texttt{c:comp}\\
\texttt{b:snomed}&\equiv&\texttt{c:snomed}\\
\texttt{b:finding}&\equiv&\texttt{c:finding}\\
\textit{p1}&\sqsubseteq_{p}&\textit{p2}
\end{eqnarray*}
\normalsize

where \footnotesize\textit{p1}=\texttt{b:ECGNormalDiag.b:component[b:P-Ax]} \normalsize and \footnotesize\textit{p2}=\linebreak\texttt{c:ECGDiagnosis.c:hasObs[c:ECGRecording].c:comp[c:P-} \texttt{Axis]}.
\normalsize

Moreover, let us suppose that system $A$ wants to send to system $B$ the following information about the ECGDiagnosis whose \textit{code} is \textit{ecg01}:
\vspace{0.3cm}

\noindent\LARGE$\sigma$\normalsize$_{code='ecg01'}$(\textit{ECGDiagnosis})= (ecg01, Normal ECG, r01)

\noindent\LARGE$\sigma$\normalsize$_{code='r01'}$(\textit{ECGRecording})= (r01, ax01, gl01)

\noindent\LARGE$\sigma$\normalsize$_{code='ax01'}$(\textit{ECGAxis})= (ax01, 27, 88, 49)

\noindent\LARGE$\sigma$\normalsize$_{code='gl01'}$(\textit{ECGGlobal}) = (gl01, 138, 390, 39, 112, 62)

\vspace{0.3cm}

Finally, assume that some of the mapping axioms between the application ontology of system A and the canonical ontology are the following:
\newcommand{\mysep}{\vspace{0.08cm}}

\scriptsize
\begin{center}
\begin{tabular*}{6in}{cc}
\texttt{a:loinc}$\equiv$\texttt{c:loinc}&
\texttt{a:snomed}$\equiv$\texttt{c:snomed}\mysep\\
\texttt{a:finding}$\equiv$\texttt{c:hasFinding}&
\texttt{a:recording}$\sqsubseteq$\texttt{c:hasObs}\mysep\\
\texttt{a:hasAxis}$\sqsubseteq$\texttt{c:comp}&
\texttt{a:value}$\equiv$\texttt{c:value}\\
\end{tabular*}
\end{center}
\normalsize


The process that needs to be carried out is composed of several steps:

\textbf{Step 1: Classification of the information in the application ontology.} In this step the information to be sent is converted into statements about individuals generated for the application ontology of system $A$. For example, the main individual \footnotesize \texttt{a:ecg01} \normalsize will be an instance of the class \footnotesize \texttt{a:ECGDiagnosis}. 
\normalsize This is a straightforward process thanks to the $\Sigma$ links created by the DB2OntoModule between the storage system of system A and its application ontology. Among others, the following OWL statements (represented as triples) will be created:

\scriptsize
\begin{center}
\begin{tabular*}{3.5in}{ll}
(\texttt{a:ecg01} \textit{\texttt{rdf:type}} \texttt{a:ECGDiagnosis})&\hspace{-2mm}(\texttt{a:r01} \textit{\texttt{a:hasAxis}} \texttt{a:ax01})\mysep\\
(\texttt{a:ecg01} \textit{\texttt{a:finding}} \texttt{a:f01})&\hspace{-2mm}(\texttt{a:ax01} \textit{\texttt{rdf:type}} \texttt{a:ECGAxis})\mysep\\
(\texttt{a:f01} \textit{\texttt{a:value}} \texttt{"Normal ECG"})&\hspace{-2mm}(\texttt{a:ax01} \textit{\texttt{a:hasP-Axis}} \texttt{a:pax01})\mysep\\
(\texttt{a:ecg01} \textit{\texttt{a:recording}} \texttt{a:r01})&\hspace{-2mm}(\texttt{a:pax01} \textit{\texttt{rdf:type}} \texttt{a:P-Axis})\mysep\\
(\texttt{a:r01} \textit{\texttt{rdf:type}} \texttt{a:ECGObservation})&\hspace{-2mm}(\texttt{a:pax01} \textit{\texttt{a:value}} \texttt{27})\\
\end{tabular*}
\end{center}
\normalsize 

%
%

\textbf{Step 2: Enrichment of the local information at the application ontology.} In this step implicit information (regarding the individuals) that can be inferred from the application ontology of system $A$ is made explicit with the help of a reasoner. For example, in this step each individual inherits a terminology code from its corresponding class:

\scriptsize
\begin{center}
\begin{tabular*}{3.5in}{ll}
(\texttt{a:ecg01} \textit{\texttt{a:loinc}} \texttt{8601-7})&(\texttt{a:ax01} \textit{\texttt{a:loinc}} \texttt{8607-4})\mysep\\
(\texttt{a:f01} \textit{\texttt{a:snomed}} \texttt{102593009})&(\texttt{a:pax01} \textit{\texttt{a:loinc}} \texttt{8626-4})\mysep\\
(\texttt{a:r01} \textit{\texttt{a:loinc}} \texttt{34534-8})&\\

\end{tabular*}
\end{center}
\normalsize


\textbf{Step 3: Classification of the information in the canonical ontology.} At this point, thanks to the equivalence, subsumption and path mappings that have been defined by the MappingModule and the help of a reasoner, the individuals are now classified as instances of the concepts of the canonical ontology. For example, given that \footnotesize \texttt{a:ECGObservation}$\equiv$\texttt{$\exists$loinc.\{34534-8\}} \normalsize and \footnotesize \texttt{c:ECGRecording}$\equiv$\texttt{$\exists$loinc.\{34534-8\}} \normalsize it is wise to think that the MappingModule will infer the equivalence mapping \footnotesize \texttt{a:ECGObservation}$\equiv$\texttt{c:ECGRecording}. \normalsize Then, as the assertional box of the application ontology of system $A$ contains the triple \footnotesize (\texttt{a:r01} \textit{\texttt{rdf:type}} \texttt{a:ECGObservation}), \normalsize the new triple \footnotesize (\texttt{a:r01} \textit{\texttt{rdf:type}} \texttt{c:ECGRecording}) \normalsize is inferred. Moreover, since triples \footnotesize (\texttt{a:r01} \textit{\texttt{rdf:type}} \texttt{a:ECGObservation}), (\texttt{a:r01} \textit{\texttt{a:hasAxis}} \texttt{a:ax01}) \normalsize and \footnotesize (\texttt{a:ax01} \textit{\texttt{a:hasP-Axis}} \texttt{a:pax01}) \normalsize exist, path rule $R1$ is fired and the triple \footnotesize (\texttt{a:r01} \textit{\texttt{c:comp}} \texttt{a:pax01}) \normalsize is generated. The remaining new triples, some of which are shown next, can be figured out accordingly.

\scriptsize
\begin{center}
\begin{tabular*}{3.5in}{ll}
(\texttt{a:ecg01} \textit{\texttt{rdf:type}} \texttt{c:ECGDiagnosis})&\hspace{-2mm}(\texttt{a:ecg01} \textit{\texttt{c:hasObs}} \texttt{a:r01})\mysep\\
(\texttt{a:ecg01} \textit{\texttt{c:loinc}} \texttt{8601-7})&\hspace{-2mm}(\texttt{a:r01} \textit{\texttt{c:comp}} \texttt{a:pax01})\mysep\\
(\texttt{a:ecg01} \textit{\texttt{c:hasFinding}} \texttt{a:f01})&\hspace{-2mm}(\texttt{a:pax01} \textit{\texttt{rdf:type}} \texttt{c:P-Axis})\mysep\\
(\texttt{a:f01} \textit{\texttt{c:snomed}} \texttt{102593009})&\hspace{-2mm}(\texttt{a:pax01} \textit{\texttt{c:value}} \texttt{27})\mysep\\
(\texttt{a:r01} \textit{\texttt{rdf:type}} \texttt{c:ECGRecording})&\\
\end{tabular*}
\end{center}
\normalsize

%

\textbf{Step 4 : Recognition at the receiver's ontology.} The triples generated
 up to this moment are sent to system B and, thanks to the ontological mappings
 defined for this ontology by the MappingModule, the individuals will be 
recognized as instances of the classes of its application ontology. For 
example, due to \footnotesize (\texttt{a:f01} \textit{\texttt{c:snomed}} \texttt{102593009}), 
\texttt{b:snomed}$\equiv$\texttt{c:snomed} \normalsize and the definition of
 class \footnotesize \texttt{b:ECGNormalFind}, \texttt{f01} \normalsize is
 classified as an individual of class \footnotesize \texttt{b:ECGNormalFind}, 
 \normalsize and then, due to the definition of class \footnotesize 
\texttt{b:ECGDiagnosis}, 
\normalsize now the main individual
\footnotesize 
\texttt{a:ecg01} \normalsize is classified as an individual of class 
\footnotesize \texttt{b:ECGNormalDiag} \normalsize:

\scriptsize
\begin{center}
\begin{tabular*}{3.5in}{ll}
\hspace{-2mm}(\texttt{a:ecg01} \textit{\texttt{b:loinc}} \texttt{8601-7})&\mysep\\
\hspace{-2mm}(\texttt{a:ecg01} \textit{\texttt{rdf:type}} \texttt{b:ECGDiagnosis})&\hspace{-4mm}(\texttt{a:ecg01} \textit{\texttt{b:component}} \texttt{a:pax01})\mysep\\
\hspace{-2mm}(\texttt{a:ecg01} \textit{\texttt{b:finding}} \texttt{a:f01})&\hspace{-4mm}(\texttt{a:pax01} \textit{\texttt{rdf:type}} \texttt{b:P-Ax})\mysep\\
\hspace{-2mm}(\texttt{a:f01} \textit{\texttt{rdf:type}} \texttt{b:ECGNormalFind})&\hspace{-4mm}(\texttt{a:pax01} \textit{\texttt{b:value}} \texttt{27})\mysep\\
\hspace{-2mm}(\texttt{a:ecg01} \textit{\texttt{rdf:type}} \texttt{b:ECGNormalDiag})&\\
\end{tabular*}
\end{center}
\normalsize

\begin{figure}
\begin{center}	
	\includegraphics[width=3.3in]{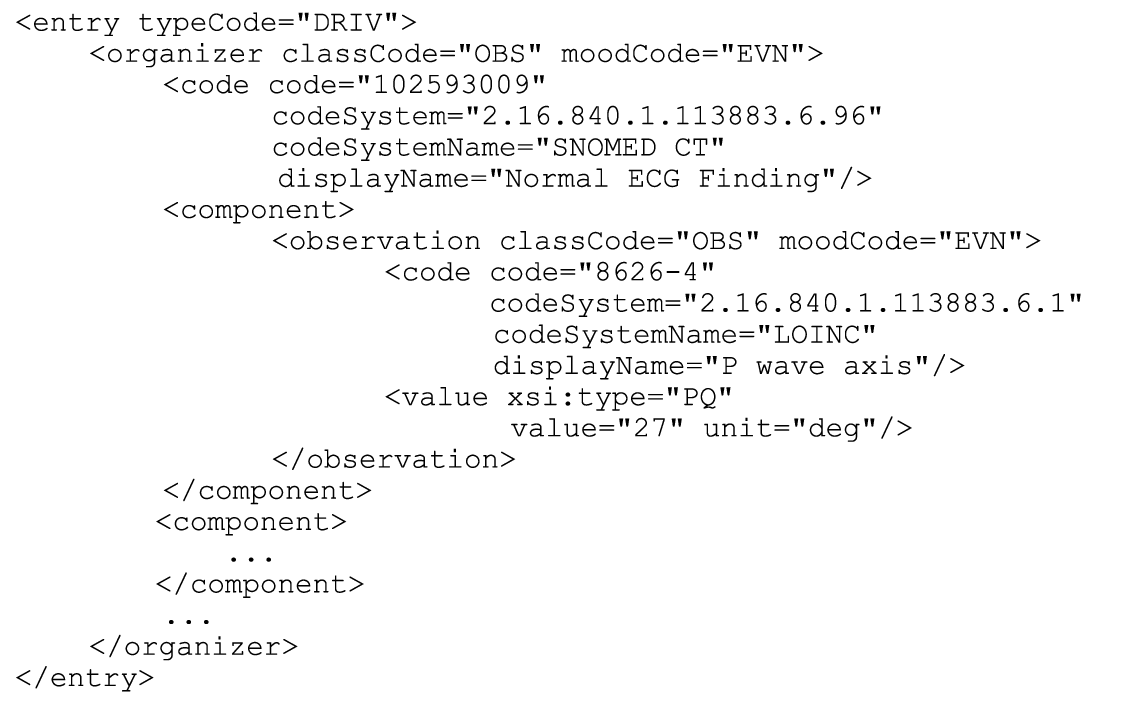}
	\caption{\label{hl7example} Excerpt of the generated HL7 entry}
\end{center}
\end{figure}

\textbf{Step 5: Storage at the receiver's system:} At this point, it is straightforward to store the information into the underlying repository of system B due to the $\Sigma$ links that indicate how to transform the collection of triples into a suitable HL7 document (see Fig.\ref{hl7example}). Notice that since the main individual \footnotesize \texttt{ecg01} \normalsize has been recognized as of class \footnotesize \texttt{b:ECGNormalDiag}, \normalsize it is possible to choose from the HL7 entry templates of system $B$ the one which represents only information about normal ECG results --despite in the sender's system there was only one table for storing all kind of ECG diagnoses. 

\section{Conclusions and discussion}\label{sec:conclusions}
We have presented a semantic-based framework which
allows the interoperability of medical diagnoses between health information systems, including those
which were not developed following EHR standards. The feasibility of the idea has been proved through an
example. To sum up, the main features of the framework presented in this paper are the following: 
(1) It is extensible to both standard and proprietary models, since any healthcare institution could create its own application ontology
and relate it to the terms of the canonical ontology via an integration mapping. Two modules are provided in order to facilitate such a technically demanding skill: one module that facilitates the task
of obtaining the definitions of the application ontology from a particular underlying
system and another module that facilitates the task of linking definitions of the application ontology to definitions of the canonical ontology; (2) It uses a formal ontology as canonical conceptual model, which
allows to focus on semantic aspects that are independent of the languages or technologies
used to describe EHRs. As a result, it is not based on peer-to-peer transformations but on the semantic acknowledgement of one instance of a class in the source ontology as instance of another class in the target ontology; (3) The features of any specific system remain unknown to the other systems in the framework. Acknowledging and using the canonical ontology as a shared model is enough; (4) Reasoning plays a major role in several parts of the framework, which decreases the need of human intervention.

However, there are still some challenges, such as those regarding scalability, that need to be addressed in order for this approach to be accepted widely. 
In the case of the DB2OntoModule, the existence of the terminology manager TM is assumed. The fact that a particular term of a database has a corresponding terminological code in the TM allows a more precise definition of that term in the application ontology. We are aware that database systems may not provide with such a set of correspondences, so syntactic and semantic similarity measures (such as Levenshtein distance\footnote{http://www.levenshtein.net/} or WordNet\footnote{wordnet.princeton.edu/}-based similarity) between the terms in the database and those in the terminologies would have to be applied in order to obtain a set of candidate codes.
Moreover, relational databases whose schema can be consulted have been chosen as underlying repositories. In the real world data can be far messier and come from unstructured or semi-structured sources. In general, the less structured the source is, the more difficult the construction of the ontology will be. In the case of unstructured sources, machine learning and text mining algorithms could be used in order to create an ontology from input documents. For semi-structured data in XML, XQuery\footnote{http://www.w3.org/TR/xquery/} and XPath\footnote{http://www.w3.org/TR/xpath20/} could be used for extraction of relevant information, and moreover, fuzzy extensions of those languages could be used to enhance that extraction. Another technique that could be applied in semi-structured sources is ILP\cite{Muggleton91}.
With respect to the core task of building an agreed canonical ontology, efforts devoted to classifications on standards (e.g. openEHR) or terminology taxonomies (e.g. SNOMED-CT) can be exploited and oriented towards the design of such an ontology. Finally, challenges concerning mappings between the application and canonical ontologies are diverse (e.g. variable granularity of the information, different types of data, etc.). As stated in section \ref{sec:mappingmodule}, extensive work has already been made on this area, so the definition of a new approach is out of the scope of this paper. However, we have presented a novel contribution concerning mapping issues: the definition of the notion of path mappings and their implementation using SWRL rules. Additionally, we suggest that specific systems publish voluntarily the integration mappings between their application ontology and the canonical ontology, so that other systems could benefit from this knowledge at the time of creating their integration mapping.



%

\section*{Acknowledgment}
The work of Idoia Berges is supported by a grant of the Basque
Government (Programa de Formaci\'on de Investigadores del Departamento de Educaci\'on,
Universidades e Investigaci\'on). This work is also supported by the Spanish Ministry of Education and Science TIN2010-21387-C02-01.

\ifCLASSOPTIONcaptionsoff
  \newpage
\fi



\bibliographystyle{IEEEtran}
\end{document}